\documentclass[10pt,twocolumn,letterpaper]{article}

\usepackage{cvpr}              

\usepackage[dvipsnames]{xcolor}

\definecolor{cvprblue}{rgb}{0.21,0.49,0.74}
\usepackage[pagebackref,breaklinks,colorlinks,citecolor=cvprblue]{hyperref}

\def\paperID{*****} 
\def\confName{CVPR}
\def\confYear{2024}

\title{Emotion Recognition Using Transformers with Masked Learning}

\author{Seongjae Min, Junseok Yang, Sangjun Lim, Junyong Lee, Sangwon Lee\\
Graduate School of Automotive Engineering\\
Kookmin University, Seoul, Korea\\
{\tt\small \{mugun19, 01079421063, sangjun7358, wnsdyd1569, sang9804\}@kookmin.ac.kr}
\and
Sejoon Lim\thanks{Sejoon Lim is the corresponding author.}\\
Department of Automobile and IT Convergence\\
Kookmin University, Seoul, Korea\\
{\tt\small lim@kookmin.ac.kr}
}

\begin{document}
\maketitle
\begin{abstract}
In recent years, deep learning has achieved innovative advancements in various fields, including the analysis of human emotions and behaviors. Initiatives such as the Affective Behavior Analysis in-the-wild (ABAW) competition have been particularly instrumental in driving research in this area by providing diverse and challenging datasets that enable precise evaluation of complex emotional states. This study leverages the Vision Transformer (ViT) and Transformer models to focus on the estimation of Valence-Arousal (VA), which signifies the positivity and intensity of emotions, recognition of various facial expressions, and detection of Action Units (AU) representing fundamental muscle movements. This approach transcends traditional Convolutional Neural Networks (CNNs) and Long Short-Term Memory (LSTM) based methods, proposing a new Transformer-based framework that maximizes the understanding of temporal and spatial features. The core contributions of this research include the introduction of a learning technique through random frame masking and the application of Focal loss adapted for imbalanced data, enhancing the accuracy and applicability of emotion and behavior analysis in real-world settings. This approach is expected to contribute to the advancement of emotional computing and deep learning methodologies. The code is here.
~\url{https://github.com/msjae/ABAW}
\end{abstract}    
\section{Introduction}
\label{sec:intro}

Recently, deep learning has undergone significant changes in various fields such as computer vision, natural language processing, and especially in analyzing human emotions and behaviors. One of the key developments in this field is the Affective Behavior Analysis in-the-wild (ABAW) competition held by Kollias et al.~\cite{kollias2019deep, kollias2019expression,kollias2019face,kollias2020analysing,kollias2021affect,kollias2021analysing,kollias2021distribution,kollias2022abaw,kollias2023abaw,kollias2023abaw2,kollias2023multi,zafeiriou2017aff,kollias20246th} These competitions facilitate research by providing diverse and challenging datasets such as AffWild2, C-EXPR-DB, and Hume-Vidmimic2, encouraging the development of models capable of accurately assessing complex emotional states. These models provide keys through Valence-Arousal (VA) estimation, facial expression recognition, and Action Unit (AU) detection, which are essential components in understanding human emotions.

In the field of emotional analysis, VA estimation provides the foundation by quantifying the positivity (Valence) and intensity (Arousal) of emotions, while facial expression recognition focuses on classifying facial expressions into distinct emotions. Furthermore, Action Unit detection emphasizes identifying the basic muscle movements that constitute these expressions, offering finer details in interpreting emotional states.

Recent studies have embraced various deep learning approaches, including Convolutional Neural Networks (CNNs) and Long Short-Term Memory (LSTM), achieving notable success. Additionally, the emergence of transformer models has introduced a new paradigm in understanding temporal and spatial features, expanding the limits of how machines can interpret human emotions and states.

This research builds on these advancements, proposing a new learning framework that utilizes temporally ordered pairs of masked features derived from facial expressions, Action Units, and valence-arousal indicators. By integrating advancements in feature extraction and sequence modeling, we aim to refine the accuracy and applicability of emotional and behavioral analysis in real-world environments and contribute to the evolving landscape of emotion computing and deep learning methodologies.

The main contributions of this study are as follows:
\begin{itemize}
\item Introduction of random frame masking learning technique: This study proposes a new learning method that improves the generalization ability of emotion recognition models by randomly masking selected frames.
\end{itemize}
\begin{itemize}
\item Application of Focal loss to imbalanced data: By using Focal loss, we have significantly improved the performance of the model in addressing the imbalance problem in facial expression recognition and Action Unit detection.
\end{itemize}

The advancement of deep learning has brought significant changes to the study of human emotional behavior as well. The Affective Behavior Analysis in-the-wild (ABAW) competition has been a tremendous contribution to driving such needed changes and pushing the field forward. ABAW provides a wide variety of datasets, including Aff-Wild2 and C-EXPR-DB, for challenges and research opportunities. In this direction, apart from the Hume-Vidmimic2 dataset, it proposes a challenge with a number of tasks.

Valence-Arousal Estimation is a type of emotional analysis that gives an emphasis on forecasting the Valence and Arousal of the persons. Valence is referred to as a characteristic of either positivity or negativity of the emotions. An increase in Valence will symbolize an increase in positive emotion, while a reduction in Valence will show negative emotions. Greater Arousal would indicate that the emotions were more actively energized, while lesser Arousal would mean that the emotions were cool and composed. Recent studies have been doing quite well with performance in ~\cite{oh2021causal} using CNNs and LSTMs. Some recent progress has been reported in the application of transformer models as well.

The task for Expression Recognition is a mutually exclusive class recognition problem. Each frame of the video should be classified to one of the defined categories: Neutral, Anger, Disgust, Fear, Happiness, Sadness, Surprise, Other. The research has been carried out using visual and audio information where there exists, to a greater extent, emotional content. References include \cite{zhou2023leveraging, zhang2023multi,zhang2023multimodal}. Nguyen et al.\cite{nguyen2023transformer} proposed to use only images, and for each of them, a feature vector is extracted using a pretrained network and then supplied to a transformer encoder.

In Action Unit Recognition, the determination of specific Action Units (AU) based on the human face's features is done in every frame of the video. It requires facial motion analysis down to the last detail. Yu et al.\cite{yu2023local} proposed a feature fusion module based on self-attention, which is responsible for integrating overall facial characteristic and relationship feature between AUs. Zhang et al.\cite{zhang2023multi} and Wang et al.~\cite{wang2023spatio} initialized a Masked Autoencoder This enabled the extraction of various general features associated with the face.
\section{Approach}
\label{sec:approach}

\begin{figure*}[h]
    \centering
    \includegraphics[width=1\linewidth]{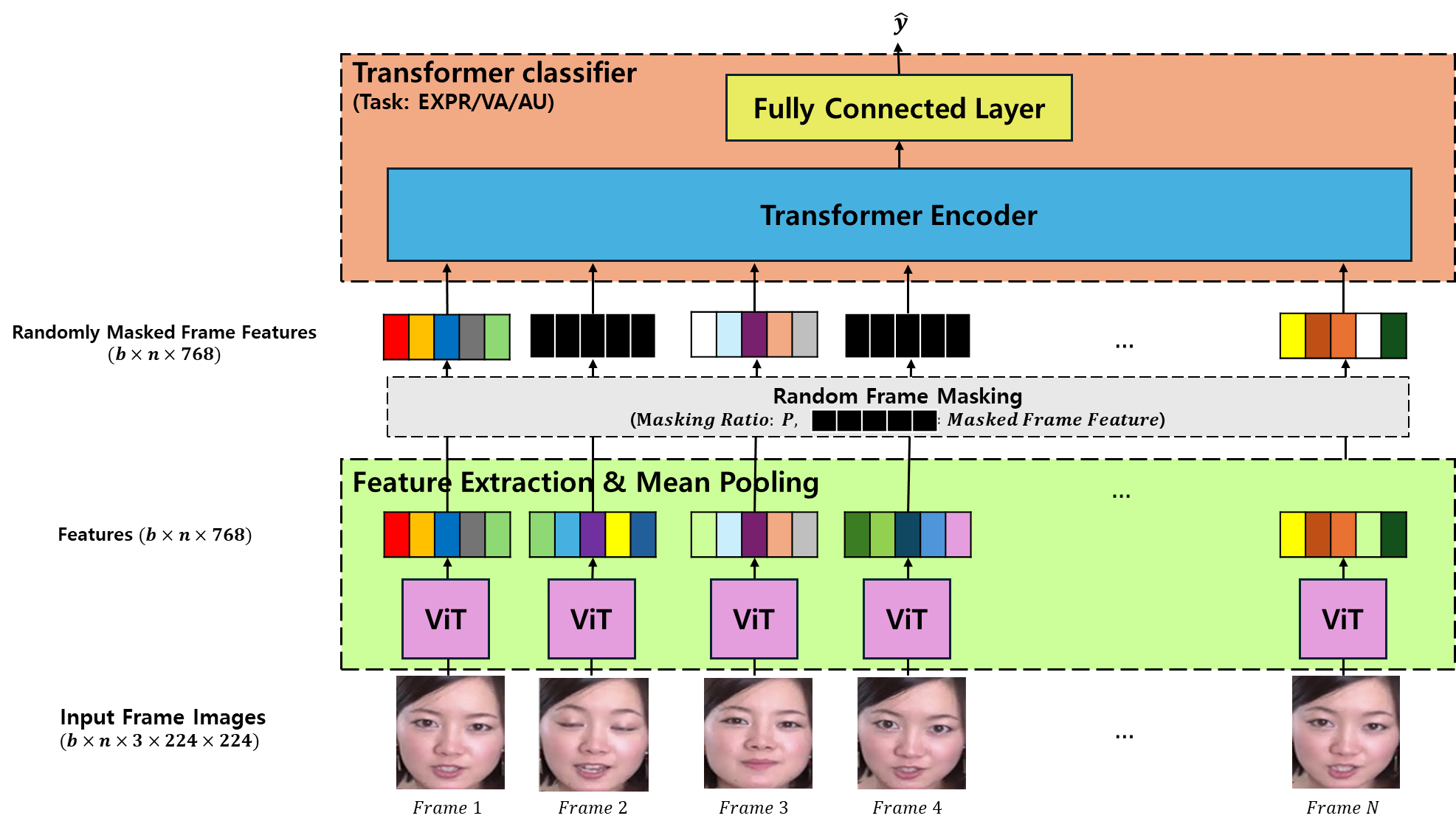}
    \caption{illustrates the comprehensive pipeline of the our model. Initially, A pretrained vision transformer individually extracts features from each input frame image (where $b$ stands for batch size, and $n$ represents sequential length), ensuring a detailed analysis of every frame. To avert the risk of overfitting, these extracted features from each frame are randomly masked. In the final step, a transformer classifier sequentially processes these randomly masked frame features to predict the outcome $\hat{y}$}
    \label{fig:pipeline}
\end{figure*}

In this paper, we suggest a network that can learn human expressions, action units (AU), valence-arousal (VA), and temporally masked features for each frame. So, the first step is the feature extraction for each of the input images. Section ~\ref{subsec:FE} details the feature extraction step. After the extraction of the features has been acquired, they are masked randomly, put together into temporal pairs, and finally input into the transformer encoder. This is followed by an FC layer to produce the final output. we describe the functioning principle of the transformer classifier module at ~\ref{subsec:TC}. Section ~\ref{subsec:LF} describes the loss function used for learning. In Figure ~\ref{fig:pipeline}, we show the schema of our whole network.

\subsection{Feature Extractor}
\label{subsec:FE}
Pretrained Vision Transformer (ViT)~\cite{dosovitskiy2020image} network in order to extract useful features Instead of using the 'cls' token from ViT's final output in the conventional manner, we apply average pooling to the output of the last layer based on the method put forth in ~\cite{beyer2022better}. This saves computational power required during training time by pre-extracting features for the Aff-Wild2 dataset. Instead, the use of a large-scale pretrained network allows for the extraction of generalized representations better adapted to the diverse contexts of the image and enhancement of its ability in both processing and analyzing the input image's complex emotional expressions and related action units.

\subsection{Transformer Classifier}
\label{subsec:TC}
Masked inputs in the Transformer model have been validated in different parts of the Transformer Classifier: GPT ~\cite{brown2020language}, Bert ~\cite{devlin2018bert}, MAE ~\cite{he2022masked} Motivated from the works discussed above, we propose to design a Transformer Classifier with features processed in the order of time and an input mask. The proposed encoder is designed to realize the self-attention mechanism that can process efficient sequences of image data. However, this approach improves to a great extent the knowledge of changes in facial expression in a temporal image sequence, something very important in the correct recognition of emotion and AU. During learning, the temporal feature pairs are given as input with a certain probability p by making them partially masked beforehand. This ensures that overfitting is totally avoided and, in turn, increases the generalization performance.

\subsection{Loss function}
\label{subsec:LF}
For AU and Expression, Focal loss ~\cite{lin2017focal}, which is strong against the imbalanced distribution of data. Focal loss performs very well while learning the model in severe class imbalanced datasets. It is defined as follows:
\begin{equation}
  L_{\text{focal}} = -\alpha (1 - p_t)^{\gamma} \log{p_t}
  \label{eq:focalloss}
\end{equation}
Here, $p_t$ denotes the predicted probability, and $\alpha$ and $\gamma$ are tuning parameters. These hyperparameters assign more importance to hard samples while reducing their importance for easier ones, so that the model gets to focus more on the part it struggles with in the learning process. This, in turn, is a performance booster for the focal loss on imbalanced datasets.

For VA measurement, CCC (Concordance Correlation Coefficient) loss was used. It is computed as follows: 
\begin{equation}
  L_{\text{ccc}} = 1 - \frac{2 \rho \sigma_{xy}}{\sigma_x^2 + \sigma_y^2 + (\bar{x} - \bar{y})^2}
  \label{eq:cccloss}
\end{equation}

Here, $\rho$ is the correlation coefficient between the two variables, and $\sigma_{xy},\sigma_x^2,\ \sigma_y^2$represent their respective averages. The CCC loss function measures the concordance between the predicted and actual values, making it a suitable loss function for predicting emotional states.
The foregoing greater importance to the difficult samples and reducing the importance of easy samples allows the model to focus more on parts that should be more concentrating in the learning process. Then, Focal loss will be the best function that will substantially improve the performance in a highly imbalanced dataset. CCC loss function, on the other hand, is best used in predicting emotional states because it gives a way that makes it possible to quantify the agreement between the predicted and the target values.
\section{Experiments}

\subsection{Experimental Setup}
In the current study, the researcher used the ImageNet21k and Aff-Wild2 datasets. ImageNet21k refers to a large-scale dataset that consists of approximately 21,000 classes and has around 14 million images, which was used in pre-training the feature extractor. The Aff-Wild2 input model was only used if the cropped image was available and was utilized for the Transformer Encoder training.

\subsection{Implementation}
The feature extractor uses ViT Base. The Transformer Classifier utilizes 8 heads, 6 layers, and a dropout rate of 0.2. The batch size is set to 512, and the temporal length is set to 100. The optimizer used is AdamW with a learning rate of 0.0001 and a fixed weight decay of 0.001. The parameters for Focal loss are set with alpha at 0.25 and gamma at 2.

\begin{table}[h]
    \centering
    \begin{tabular}{lccr}
    \hline
             Challenge & Metric    & Method   & Result         \\ \hline
    VA        & CCC       & Ours     & 0.32           \\
              &           &          & (CCCv:0.23, CCCa:0.41) \\
              &           & Baseline & 0.22           \\
              &           &          & (CCCv:0.24, CCCa:0.20) \\ \hline
    EXPR      & F1-Score  & Ours     & 0.29           \\
              &           & Baseline & 0.25           \\ \hline
    AU        & F1-Score  & Ours     & 0.40           \\
              &           & Baseline & 0.39           \\ \hline
    \end{tabular}
    \caption{Results on Validation set of Aff-Wild2}
    \label{tab:performance_comparison}
\end{table}

\subsection{Results}
~\ref{tab:performance_comparison} presents the test results of our methodology on the validation set for VA, EXPR, and AU, compared with the Baseline. For VA Estimation, we used the average of the Concordance Correlation Coefficients for Valence and Arousal. For EXPR Recognition and AU Detection, we used the F1 Score as the evaluation metric for each, respectively.

{
    \small
    \bibliographystyle{ieeenat_fullname}
    \bibliography{main}
}


\end{document}